\documentclass[letterpaper, 10 pt, conference]{ieeeconf}  

\IEEEoverridecommandlockouts                              

\usepackage{graphics} 

\usepackage{graphicx}
\usepackage{caption}
\usepackage{subcaption}
\usepackage{color,soul}
\usepackage{amsmath}
\usepackage{siunitx}
\usepackage{epstopdf}

\usepackage{epsfig} 
\usepackage{mathptmx} 
\usepackage{times} 
\usepackage{amsmath} 
\usepackage{amssymb}  

\usepackage{breqn}

\DeclareMathAlphabet\mathbfcal{OMS}{cmsy}{b}{n}

\title{\LARGE \bf
Kinematics and Dynamic Modeling of a Planar Hydraulic Elastomer Actuator
}

\author{Mahdi Momeni Kelageri, Mikko Heikkil{\"a}, Jarno Jokinen, Matti Linjama, Reza Ghabcheloo$^{1}$
\thanks{*This work was supported by Academy of Finland under the ActiveFit project (No. 295817)}
\thanks{$^{1}$Authors are with Tampere University of Technology, Tampere, Finland
        {\tt\small \{mahdi.momenikelageri, mikko.heikkila, matti.linjama, reza.ghabcheloo\}@tut.fi}}%
}

\begin{document}

\maketitle
\thispagestyle{empty}
\pagestyle{empty}


\begin{abstract}

This paper presents modeling of a compliant 2D manipulator, a so called soft hydraulic/fluidic elastomer actuator. Our focus is on fiber-Reinforced Fluidic Elastomer Actuators (RFEA) driven by a constant pressure hydraulic supply and modulated on/off valves. We present a model that not only provides the dynamics behavior of the system but also the kinematics of the actuator. In addition to that, the relation between the applied hydraulic pressure and the bending angle of the soft actuator and thus, its tip position is formulated in a systematic way. We also present a steady state model that calculates the bending angle given the fluid pressure which can be beneficial to find out the initial values of the parameters during the system identification process. Our experimental results verify and validate the performance of the proposed modeling approach both in transition and steady states. Due to its inherent simplicity, this model shall also be used in real-time control of the soft actuators.

\end{abstract}


\section{INTRODUCTION}

Biology has long been an important source of inspiration for the engineers in order to make ever-more capable machines. Of noticeable features exploited from biological systems is softness and body compliance which tend to seek simplicity in interaction of such systems with their environment. Several of the lessons learned from studying biological systems are now culminating in the definition of a new class of machines that is referred to as soft robots \cite{Daniela2015_Nature}. Soft material actuators are made of deformable materials which means that, theoretically, the actuator tip can attain every point in $3$D workspace with an infinite number of configurations.

To improve the performance of a system, its behavior has to be modeled more accurately. Furthermore, the possibility to calculate the dynamic model in real-time is required for control algorithms. Soft material actuators, in particular, can potentially undergo large deformations which turns them to somehow complex systems which are not straightforward to acceptably accurately model.

\begin{figure}
    \centering
    \begin{subfigure}[t]{0.23\textwidth}
        \includegraphics[width=\textwidth]{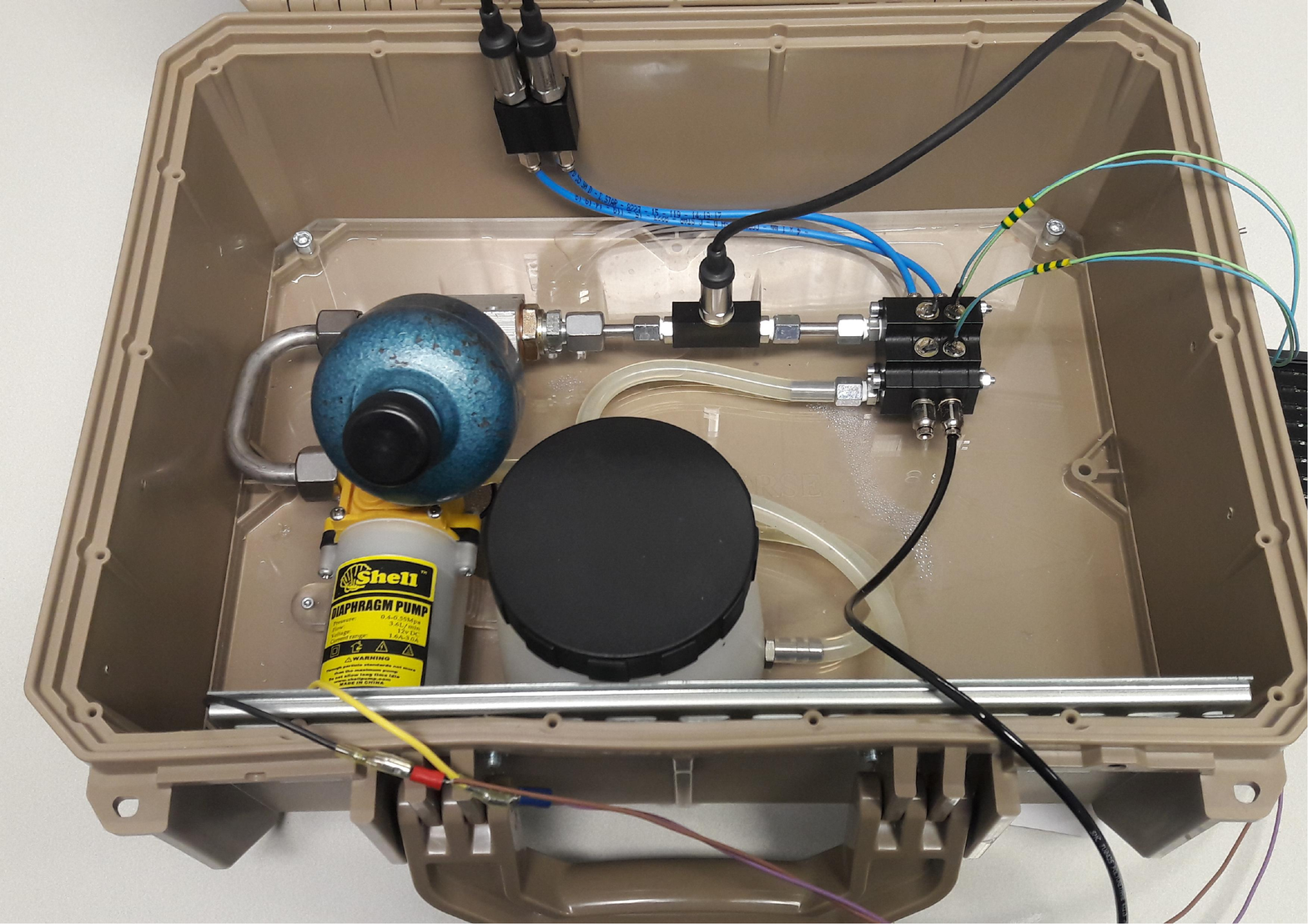}
        \caption{test system}
        \label{fig:DHTestSystem}
    \end{subfigure}
    ~ 
    \begin{subfigure}[t]{0.23\textwidth}
        \includegraphics[width=\textwidth]{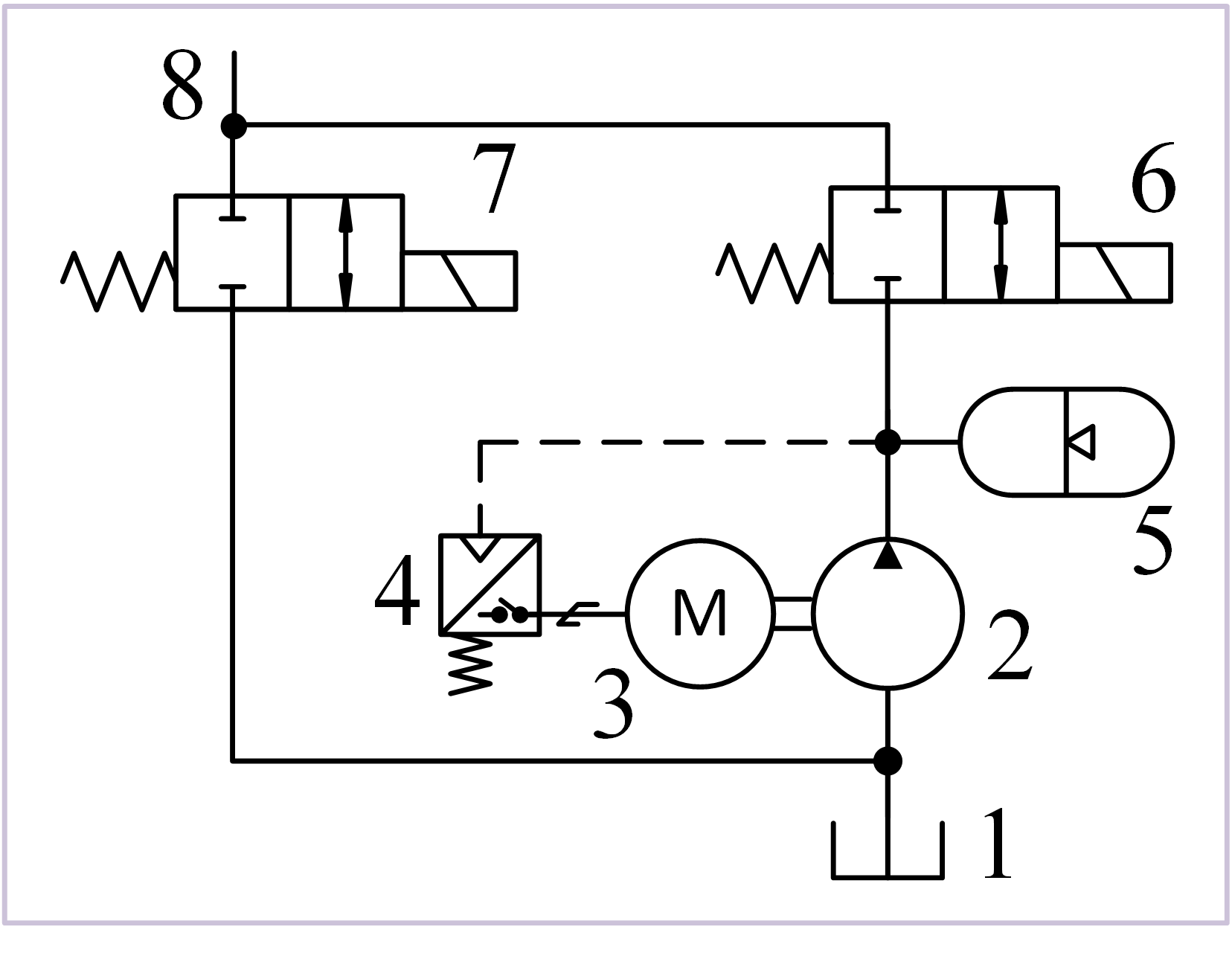}
        \caption{schematic diagram}
        \label{fig:DHSchematicDiagram}
    \end{subfigure}
    ~ 
    \caption{Digital hydraulic drive system for controlling a soft actuator}
    \label{fig:DigitalHydraulic}
\end{figure}


\subsection{Prior Work}

The majority of works in modeling of soft robots deal with the kinematics of the system. Not many works have been presented on dynamic modeling of the system. There are also some works on modeling of soft robots using finite elements method (FEM). To the best of our knowledge, there is no work incorporating both kinematics and dynamics models at the same time, which is also fast enough to be implemented in real-time control applications. In this section we will briefly review some of the most important prior works. 

In \cite{Hannan2003} the authors develop a continuum kinematics for an elephant trunk and demonstrate how it can be used in obstacle avoidance. In \cite{Jones2006} a kinematic algorithm for controlling the shape of multi-segment continuum manipulators is presented. Despite numerous continuum manipulator designs, it is often possible to find out their kinematics based on a piece-wise constant curvature (PCC) model \cite{Webster2010}. However, there are two important, somehow related, issues with PCC models that can seriously influence on its accuracy and restrict its application: 1) when the deformation is large, and 2) the deformation is not according to the PCC assumption in all soft material actuators. For this reason, the authors in \cite{Mahl2013} developed general variable curvature continuum kinematics.

One common approach to shape estimation in continuum robots is measuring strain along the manipulator axis \cite{Leleu2001, Lyshevski2003} and applying dynamical models of the manipulator in order to predict the curvature of the manipulator based on measured strain. Another method for shape estimation is using fiber optic sensors along the body. This method, however, suffers from propagation losses when they are bent \cite{Miller2004}. In \cite{Hannan2005} the authors use vision system to estimate the curvature. Vision systems are usually based on PCC assumptions that can be inaccurate if large deformation and gravitational loading effects are present. In \cite{Trivedi2014} the authors present three different methods of shape estimation based on geometrically exact mechanical model which are: $1)$ a load cell mounted at the base of the manipulator $2)$ using cable encoders running through the length of the actuator and $3)$ using inclinometers mounted at the end of each section of the manipulator.

Dynamic modeling of soft material actuators capable of applying to real-time control is a very complex task and thus, most papers in the field treat the control problem from the kinematic point of view. The dynamic modeling of extensible soft elastomer actuators is an important open research field. In \cite{Takegaki1981} a dynamic model for hyper-redundant structure as an infinite degree of freedom continuum model is proposed. In \cite{Boyer2006} the authors presented dynamic model for an eel-like robot. The authors in \cite{Ivanescu2005} represent a dynamic model for an extensible tentacle arm. However, their model is derived based on the restrictive assumption that the arm doesn't bend past a small-strain region, where linear stress-strain are obeyd and permanent deformation does not exist. A three-dimensional dynamic model is presented in \cite{Mochiyama2003} for an inextensible continuum manipulator based on the assumption that the continuum robot is a combination of rigid slices. The dynamic model is then derived by obtaining the limit of serial rigid chain model as the degree of freedom goes to infinity. Further, the authors in \cite{Tatlicioglu2007} extend the model to include the extensible continuum manipulator. However, they work applies to the continuum manipulator with no torsional effect. Besides, there is no comparison between the simulation results and the experimental measurements in order for the model to be evaluated.

Another approach for modeling soft actuators is a common numerical method for solving engineering problems, i.e., Finite Element Method (FEM). However, the design process of a rubber actuator is difficult because of large deformation and material nonlinearity. One of the FEM advantages is its capability for taking into account nonlinearities \cite{Nakamatsu2008} caused by material, large deformation or contact. Typically, analysis of soft material actuators have limited to small strains and linear material behaviour \cite{Moseley2016}. Usage of three-dimensional solid elements allows modelling three-dimensional deformation of the actuator, which can be used in the optimization of the actuator cross-section, for example \cite{Elsayed2014}. Non-linear FEM is an effective method for design process \cite{Suzumori2007}. The disadvantage of the solid FEM model is the large number of solvable parameters, which increases analysis time. As a matter of fact, the main benefits of FEM modeling lies in design process and they are not very useful for real-time control applications.


\subsection{Contributions}

In this paper, we present an approach for kinematics and dynamic modeling of Reinforced Fluidic Elastomer Actuator (RFEA) taking advantage of the revolute spring-damper actuators and Lagrangian equations. Closest work to our paper is a quite recent paper \cite{Wang2017}, where they model a soft actuator using series of lines connected with viscoelastic joints. They assume that the applied torque is simply a linear function of applied pressure, i.e., $\tau_i(P) = \alpha_i P$ and $\alpha_i$ to be identified. They generally develop their work based on this to be identified equation. Whereas in our work, we derive the equations based on system parameters which gives a better insight to the idea of modeling. Our equations agree with and validate some of the assumptions in \cite{Wang2017}. Furthermore, we present a steady state model which acceptably accurately provides the bending angle (and thus the kinematics) of the actuator given the physical values of the actuator. This steady state model is also beneficial for identifying the unknown parameters. More specifically, this paper contributes the following:

\begin{itemize}

\item An analytical formulation for dynamic and kinematics modeling of RFEA based on Lagrange equations. This model not only provides the dynamic behavior of the system but also the kinematics of the elastomer actuator, given the applied pressure as the input to the system;

\item Presenting a novel and acceptably accurate steady state model which relates the bending angle to the inner/outer tube diameter and the fluid pressure. The model is used during identification of the dynamic model's parameter;

\item Investigation of the impact of the number of degrees of freedom on the proposed model;


\item Extensive experiments for validation of the model.

\end{itemize}


\begin{figure}[t]
\centering
\includegraphics[trim={0 0 0 0},clip, width=3.0 cm]{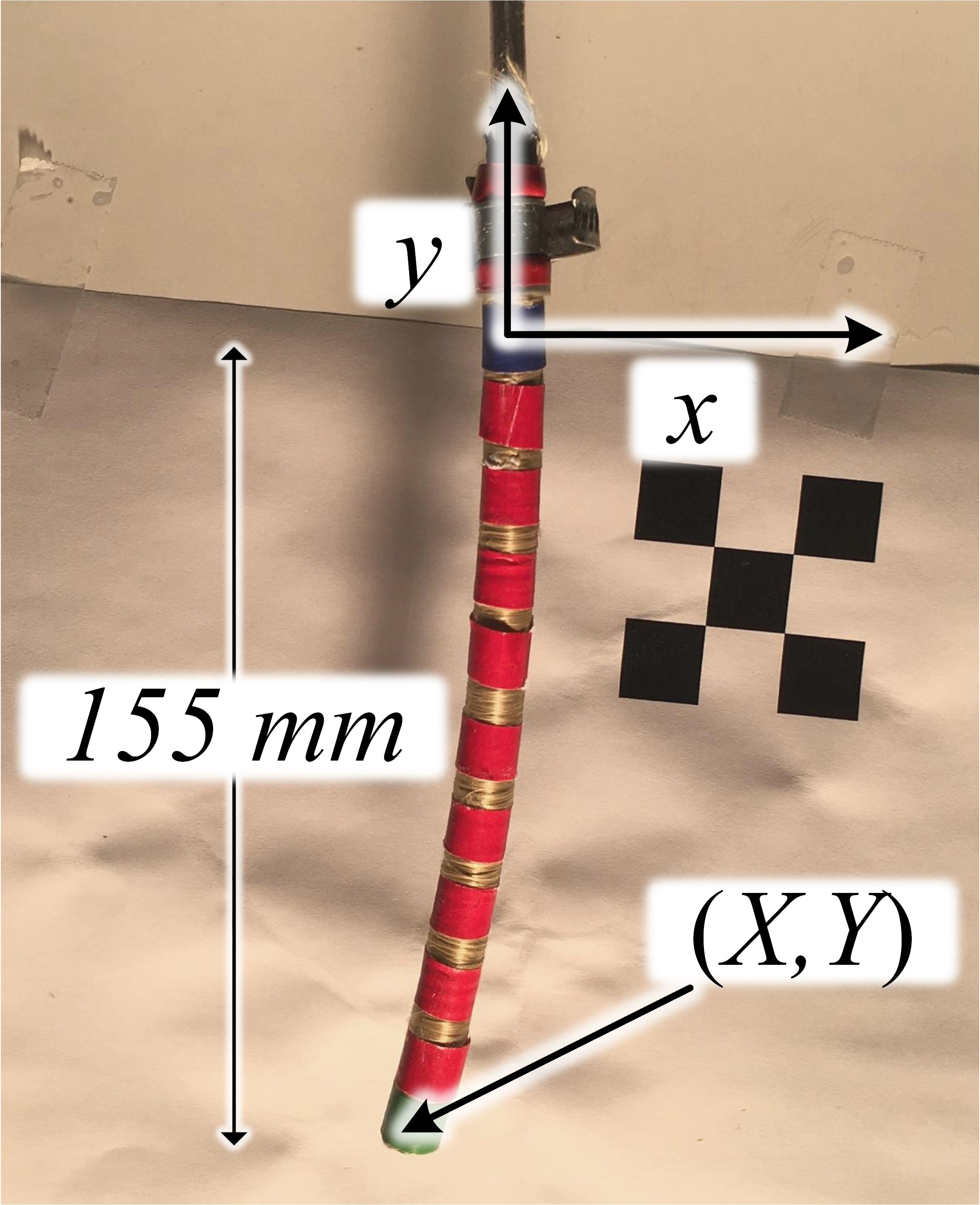}
\caption{Experimental Setup}
\label{fig:ExpSetup}
\end{figure}

\section{System Description}

The hydraulic elastomer actuator system studied in this paper is composed of: $A)$ a soft and compliant fiber-reinforced one-directional elastomer tube (outer radius $r_o = 8$mm, inner radius $r_i = 6$ mm, length $L = 155$ mm) $B)$ digital hydraulic drive system, $C)$ vision system used for model verification and identification, Fig.~\ref{fig:DigitalHydraulic} and Fig.~\ref{fig:ExpSetup}, and $D)$ control hardware. We will next describe these elements:

\subsection{Elastomer Actuator}

\begin{figure}[t]
\centering
\includegraphics[trim={0 0 0 0},clip, width=7.0 cm]{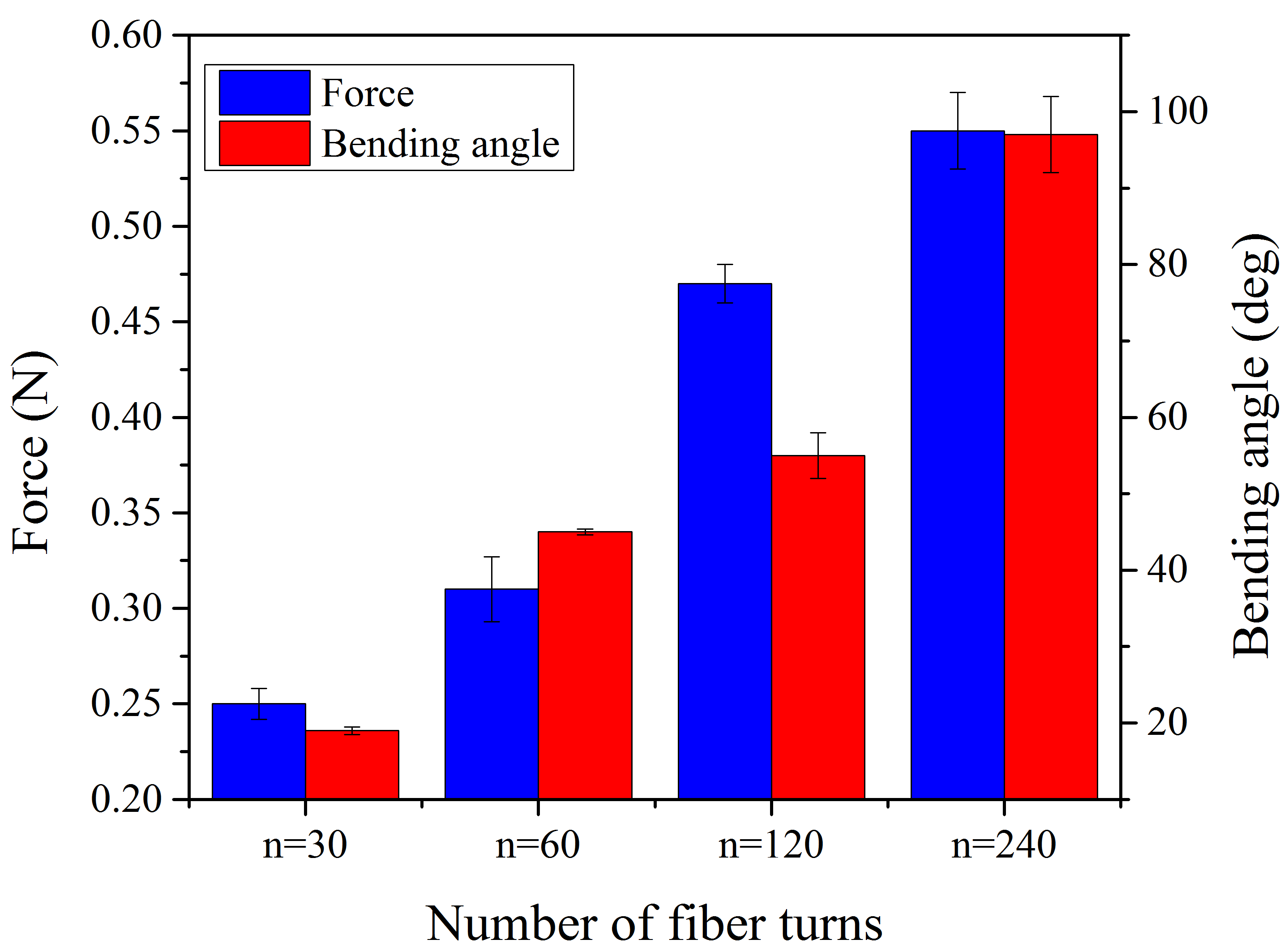}
\caption{The effect of number of turns on actuation performance of PDMS at 5 KPa}
\label{fig:ActuationPerformance-Turns}
\end{figure}

For satisfying the research requirements of this work, the actuator was designed with a cylindrical shape, as shown in Fig. ~\ref{fig:ExpSetup}, and made of polydimethylosiloxane (PDMS). The number of turns of fibers was set to $240$ as it gave the highest force output as well as bending angle, Fig.~\ref{fig:ActuationPerformance-Turns} \cite{Minna}. Bending principle is based on asymmetric structure and expansion in the direction of the lowest modulus by pressurizing or depressurizing internal fluid which induces stress in elastomer. In this work, the actuator's workspace is constrained to X-Y plane while it is only capable of one-directional bending.

\subsection{Digital Hydraulic Drive}

A digital hydraulic drive system is used to control the RFEA. The test system is shown in Fig.~\ref{fig:DHTestSystem} while the corresponding hydraulic diagram is depicted in Fig.~\ref{fig:DHSchematicDiagram}: The size of the tank 1 is $0.6$ l and it is connected to the diaphragm pump 2. The pump is then connected to a $12$ V DC motor 3 having a maximum power of $36$ W. The maximum flow rate of the hydraulic power unit 2 is $3.6$ l/min whereas the maximum system pressure is limited to about $600$ kPa by the pressure switch 4. The hydro-pneumatic accumulator 5 is attached to the supply line to store the hydraulic energy. The fluid volume in the actuator port 8 can be increased by opening the high-pressure valve 6. On the other hand, opening the low-pressure valve 7 decreases the actuator fluid volume as the flow direction is towards the tank. The orifice diameter of these on/off valves is $1$ mm. Water is used as the hydraulic medium in the drive system.

\subsection{Vision System}

To verify the simulation results, a GoPro camera has been used in order to find out the curvature/bending angle of the manipulator and localize the tip offline. The resolution and the frame per seconds (FPS) of the camera is set to $720\times1280$ and $240$ respectively, while MATLAB Image Processing and Computer Vision Toolboxes have been used for tip localization and curvature estimation. The camera is also calibrated at the initialization time before the experiment.

\subsection{Control Hardware}

A dSPACE DS$1006$ Processor Board is utilized to run pressure control algorithms of the test system. Valve control commands are executed using DS$4004$ Digital I/O Board while the hydraulic pressure in the actuator is measured using DS$2003$ A/D board. The pressure data is then recorded in a database to be used for verification of the model.


\section{Modeling}

\begin{figure}[t]
    \centering
        \includegraphics[trim={0 0 0 0},clip, width=8.0 cm]{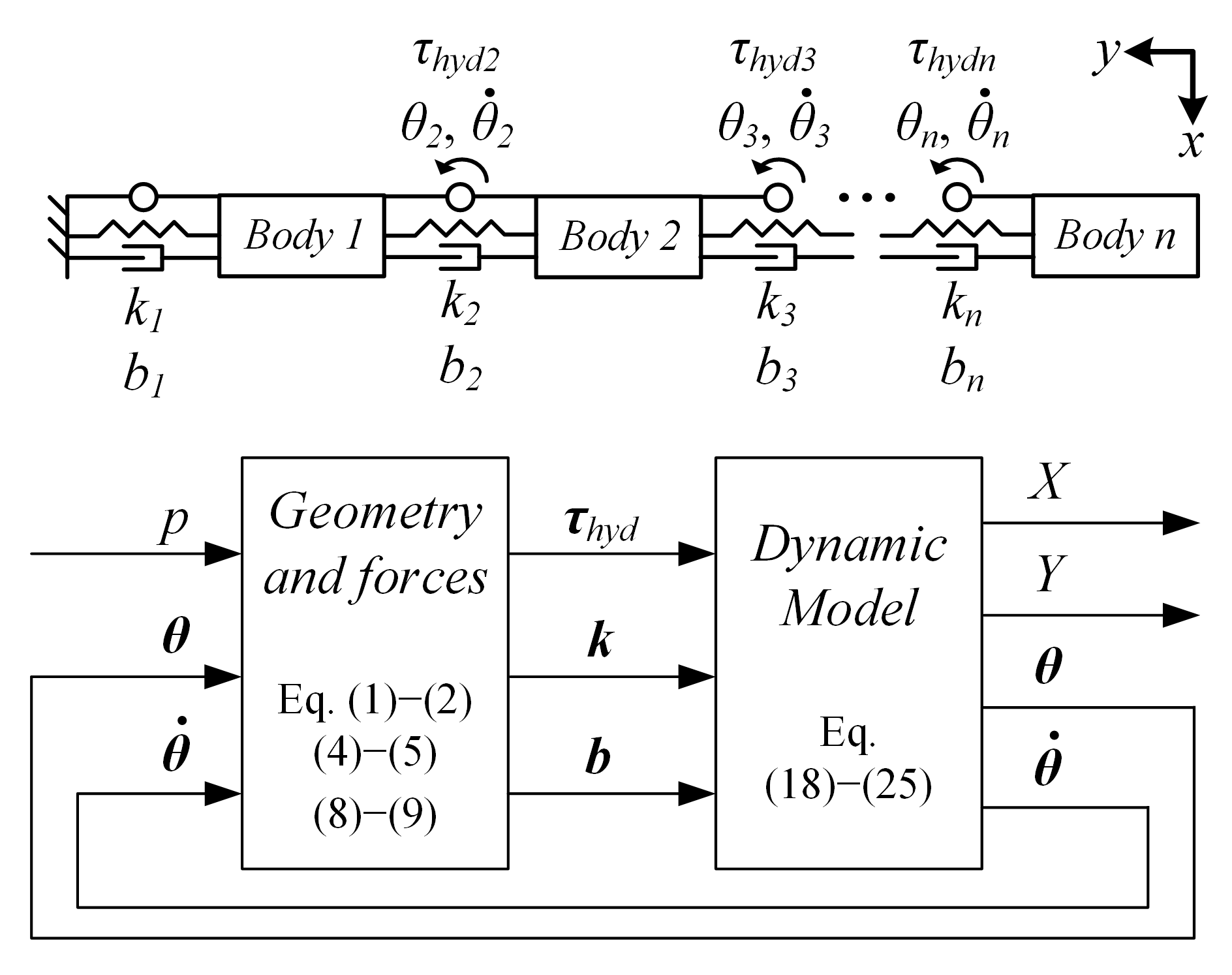}
        \caption{Modeled system}
        \label{fig:ModelDiagram}
\end{figure}

In this section, we derive the equations for the complete representation of the kinematics and dynamics behavior of the RFEA. Our goal is to derive the bending angle and tip position in a closed form equation given the physical characteristics of the actuator, i.e., inner radius $r_i$, outer radius $r_o$, actuator's length $l$, and fluid pressure $p_{hyd}$. Fig. \ref{fig:ModelDiagram} shows the overal system including the hydraulic drive system and the segmented elastomer actuator, while Fig. \ref{fig:RSDAModel} shows the schematic of the hydraulic system in more details. It should be noted that modeling the hydraulic drive system is out of the scope of this paper. We will eventually verify the model both in simulation and experimentally and compare the results. The following assumptions are valid in this work:

\begin{itemize}

\item \textit{Assumption 1:} The RFEA is moving freely in $2$D planar surface, i.e., it does not collide with an obstacle in its workspace nor it is carrying a payload.

\item \textit{Assumption 2:} The fibers are inextensible enough and prevent the actuator to radially expand. As a matter of fact, the outer radius of the actuator is assumed to be constant when the tube is pressurized.

\item \textit{Assumption 3:} The tube is made of material which has bulk modulus of about $2$ GPa. The tube, therefore, is assumed to be incompressible.

\item \textit{Assumption 4:} The material volume, $V_{rubber}$ is constant.

\item \textit{Assumption 5:} For simplicity, it is assumed that the RFEA is made of uniformly distributed material. 

\end{itemize}


\subsection{Soft Actuator Modeling}

\begin{figure}[t]
    \centering
        \includegraphics[trim={0 0 0 0},clip, width=8.0 cm]{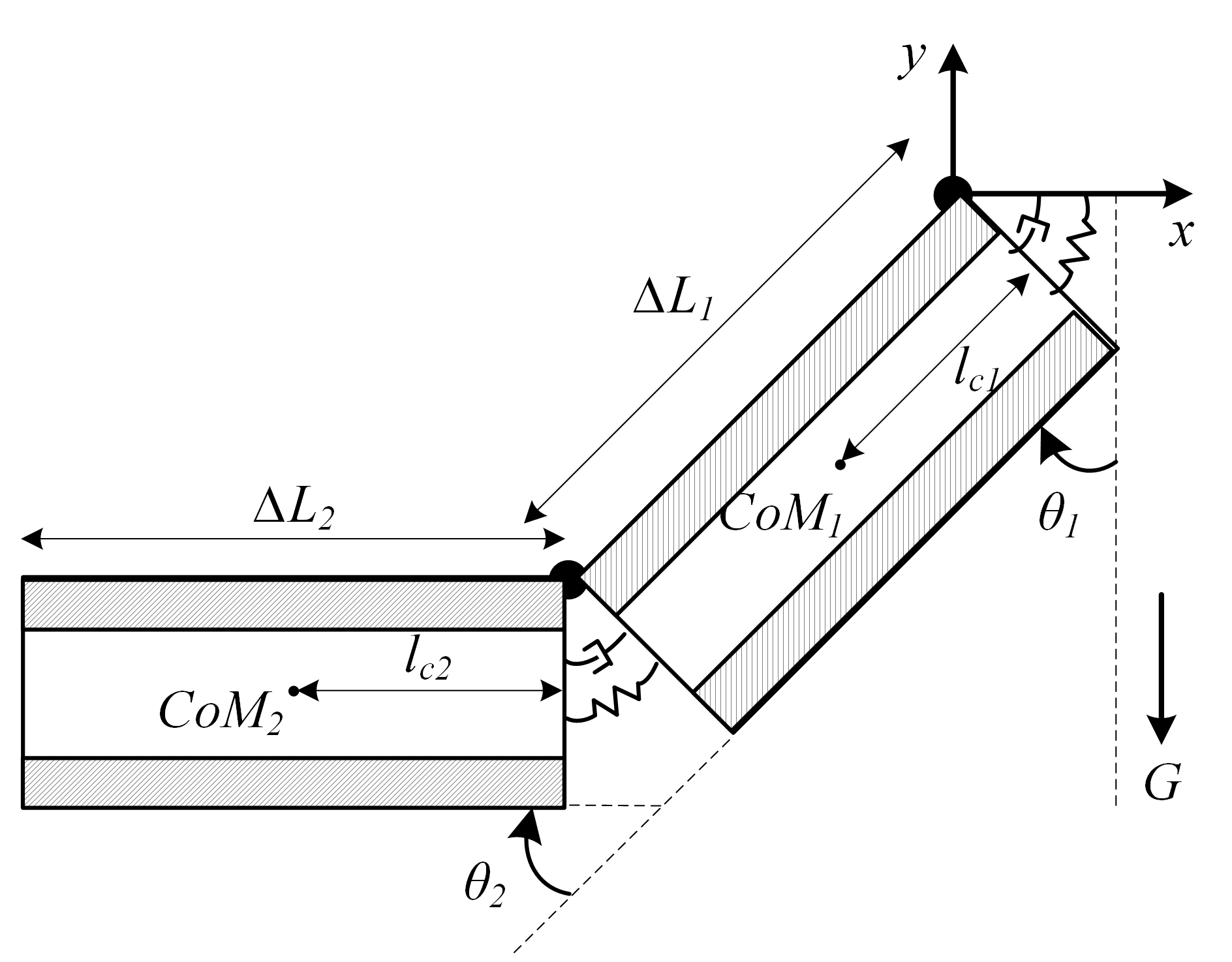}
        \caption{Mass-spring-damper system constrained with a revolute joint}
        \label{fig:RSDAModel}
\end{figure}

The main idea behind modeling the RFEA comes from the conventional mass-spring-damper system. That is to say, in order to find the dynamic and kinematics model of the RFEA, it is considered as a multibody system composed of $n$ mass-spring-damper system. However, since the actuator is a one-directional bending type, the mass-spring-damper system is constrained with a revolute joint to represent the inextensible side of the tube, a so called Revolute Spring Damper Actuator (RSDA), while the other free (unconstrained) side represents the extensible side of the soft actuator, as shown in Fig. \ref{fig:RSDAModel}. Furthermore, based on assumption 5, it is reasonable to consider the tube as a combination of equi-length bodies. In this case, we can write: $\Delta L = \frac{L}{n}$, where $L$ is the length of the tube and $n$ is the number of segments (bodies). We will call $n$ as the number of Degrees of Freedom ($n$-DoF).

We will start by modeling the soft material actuator using Lagrange equations. We will also consider the inner radius $r_i$, fluid volume $V_{seg}$, and the fluid pressure $p_{hyd}$, and thus the acting force/torque, as time dependent parameters and include their instantaneous values which contribute in model accuracy.

\subsubsection{Instantaneous Inner Radius}

The inner radius of the tube can be calculated according to the corresponding angle between the two adjacent RSDA body in the proposed multibody system. Assuming that in the fiber reinforced elastomer actuators the outer radius remains constant and also the material is incompressible, we can write:

\begin{equation}
\label{eq:DynInnerRadius}
\frac{dr_{i_j}}{dt} = \frac{r_o^2 - r_{i_j}^2}{2(s_j + L_0) r_{i_j}}\frac{ds_j}{dt}
\end{equation}

\noindent where $r_i$, $r_o$, $s$, and $L_0$ are the tube's inner radius, outer radius, elongation arc due to bending, and the initial length of the tube segment. Besides, $j = 1, \cdots, n$ is the index of segments considered for modeling the tube.\\

\subsubsection{Volume}

The volume inside the tube's $j^{th}$ segment, $V_{{seg}_j}$, is calculated according to the following equation:

\begin{equation}
\label{eq:SegmentVolume}
\frac{dV_{{seg}_j}}{dt} = \pi r_o r_i (r_i \frac{d\theta_j}{dt} + 2\theta_j \frac{dr_i}{dt})
\end{equation}

\noindent where $r_o$, the outer radius is constant and $\theta_j$ is the angle between the two adjacent segment. It should also be noted that the bending $\theta$ is the sum of the angles between the segments, i.e.;

\begin{equation}
\label{eq:BendingAngleSeg}
\theta = \sum_{j = 1}^n \theta_j
\end{equation}

\subsubsection{Acting Forces}

The forces acting on the elastomer tube are the hydraulic force generated by the fluid pressure, spring force generated by the strain (bending) and damping forces caused by friction.

The hydraulic pressure, $p_{hyd}$, is supposed to be uniformly distributed along the tube and thus, is identical for each segment. The axial hydraulic force, $F_{hyd}$, is calculated as follows:

\begin{equation}
\label{eq:HydraulicForce}
F_{{hyd}_j} = \pi r_{i_j}^2 p_{hyd}
\end{equation}

\noindent and thus, the hydraulic torque applied to each joint in the RSDA multibody system becomes:

\begin{dmath}
\label{eq:HydraulicTorque}
\tau_{{hyd}_j} = r_{hyd} F_{{hyd}_j} = p_{hyd} \pi r_{i_j}^2 r_{hyd}
\end{dmath}

\begin{figure}[t]
    \centering
        \includegraphics[trim={0 0 0 0},clip, width=8.0 cm]{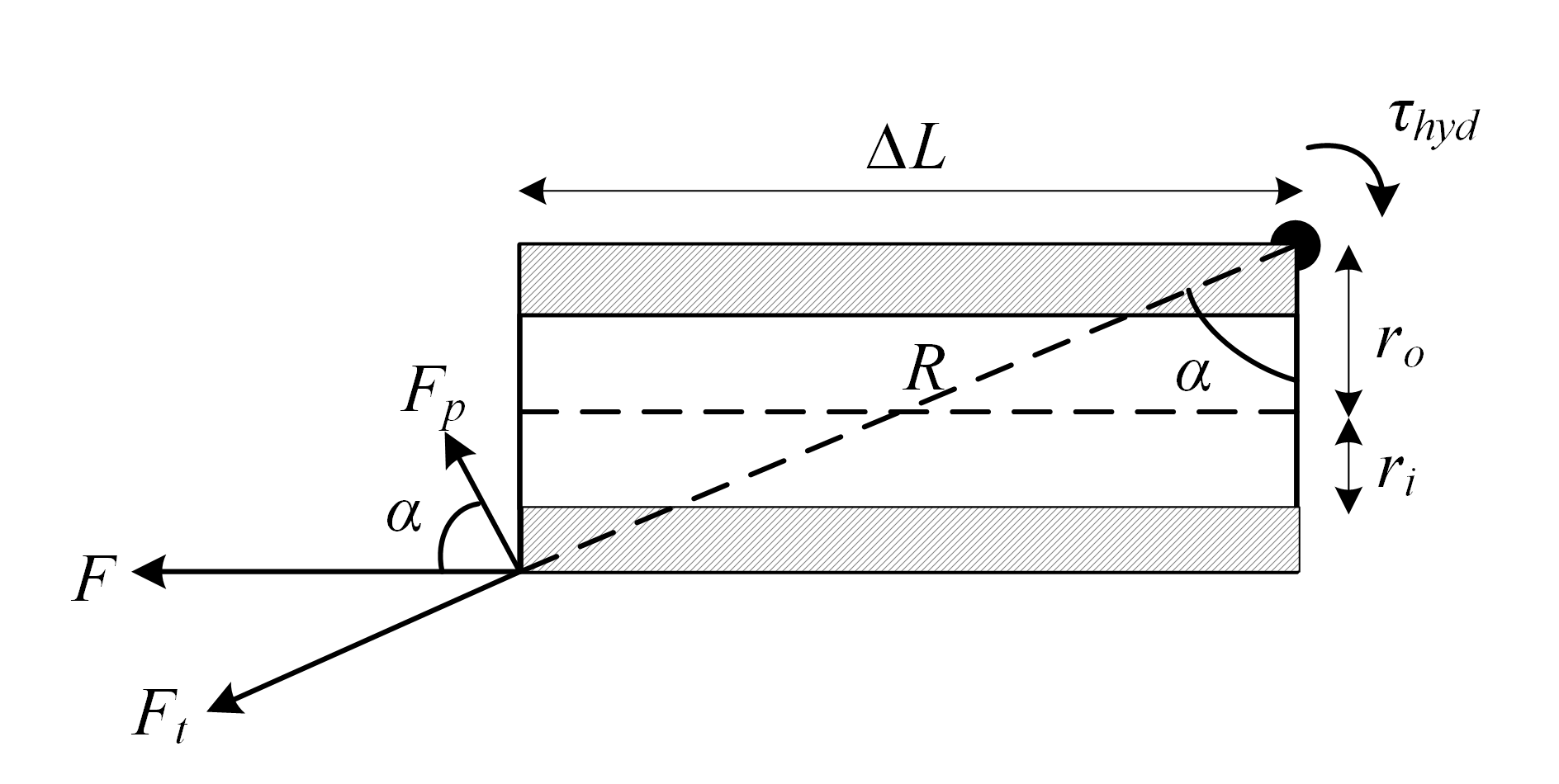}
        \caption{A segment of the tube}
        \label{fig:RSDASingleBody}
\end{figure}

\noindent where $r_{hyd}$ is the effective position vector from where the hydraulic force is applied to the system, and will be identified during system identification. However, if the number of segments in the proposed multibody system is limited, i.e., the angle $\alpha$ in Fig. \ref{fig:RSDASingleBody} is not negligible any more, then:

\begin{equation}
\label{eq:HydTorqueAlpha}
\tau_{{hyd}_j} = p_{hyd} A_j R\cos\alpha_j
\end{equation}

\noindent where:

\begin{equation}
\label{eq:AlphaR}
\begin{cases}
\alpha = \tan^{-1}\frac{\Delta L}{r_o}\\
R = \sqrt{r_o^2 + \Delta L^2}
\end{cases}
\end{equation}\\

\noindent where $\Delta L$ is the length of each body in the RSDA multibody system. It may also worth noting that our experimental results verify that the Eq. \ref{eq:HydraulicTorque} produces quite acceptable results and Eq. \ref{eq:HydTorqueAlpha} may not be really needed in practice.

To capture nonlinear mechanical behavior of the actuator material and its effect on springer and damper coefficients, we have considered that the spring and damper coefficients are changing with respect to the inner radius of the tube. Spring coefficients are, therefore, defined by:

\begin{equation}
\label{eq:SpringCoefficient}
k_j = r_{i_j} m_k + k_0
\end{equation}

\noindent where $m_k$ is a coefficient that its value is estimated through system identification. Eq.~\ref{eq:SpringCoefficient} reflects the fact that as the inner radius of the tube increases (tube is stretched), material gets less stiff (refer to the sections on experimental validation/discussion and conclusion).

The damper coefficients are also calculated by:

\begin{equation}
\label{eq:DamperCoefficient}
b_j=
\begin{cases}
r_{i_j} m_{b_{pos}} + b_{0_{pos}}	& 	: \frac{d\theta}{dt} \ge 0 \\[0.15cm]
r_{i_j} m_{b_{neg}} + b_{0_{neg}}	& 	: \frac{d\theta}{dt} < 0
\end{cases}
\end{equation}

which also includes the direction dependency of damping effect and $m_b$ is a coefficient that its value is estimated through system identification.

Even though the tube is made of uniformly distributed material, the spring/damper coefficients are different for each segment as the instantaneous inner radius is not identical. However, $k_0$, $m_k$, $b_{0_{pos}}$, $b_{0_{neg}}$, $m_{b_{pos}}$ and $m_{b_{neg}}$ are supposed to be identical for each segment.

\subsubsection{Steady State Formulation}

Assuming the proposed multibody system while the gravity is zero, in steady state the torques applied to the tube are the fluid pressure torque $\tau_{hyd}$ and the spring torque $\tau_{spring}$. The equilibrium equation for the torques can, therefore, be written as:

\begin{equation}
\label{eq:TorqueEquilibrium}
r_{hyd} F_{hyd} = r_{o} F_{spring} = \tau_{ss} \Rightarrow r_{hyd}\pi r_{i_{ss}} ^2 p_{hyd} - k_{ss}\theta_{ss} = 0
\end{equation}

\noindent where $r_{i_{ss}}$, $\theta_{ss}$ , and $\tau_{i_{ss}}$ are the inner radius of the tube, bending angle, Eq. \ref{eq:BendingAngleSeg}, and the torque in steady state, and $k_{ss} = \frac{k_j}{n}$ where $k_j$ is the coefficients in Eq. \ref{eq:SpringCoefficient}. The bending angle, therefore, can be calculated as:

\begin{equation}
\label{eq:BendingAngle}
\theta_{ss} = \frac{r_{hyd}\pi r_{i_{ss}} ^2 p_{hyd}}{k_{ss}}
\end{equation}

We assume that the revolute joints, approximately, form an arc of a circle while bent (this is a local assumption and not for the whole tube). Each segment's elongation $s$ (displacement between the center of masses of adjacent bodies), therefore, can be calculated as:

\begin{equation}
\label{eq:Elongation}
s = r_o \frac{r_{hyd}\pi r_{i_{ss}} ^2 p_{hyd}}{k_{ss}}
\end{equation}

The inner radius of the tube changes as the fluid pressure changes and so does the bending angle. Considering the assumptions of the modeling, we can write:

\begin{equation}
\label{eq:Volume}
V = constant \Rightarrow L_0 \pi (r_o^2 - r_{{i}_0}^2) = \pi (L_0 + s)(r_o^2 - r_{i_{ss}}^2)
\end{equation}

\noindent where $r_{i_0}$ is the initial inner radius of the tube, i.e., while the tube is unpressurized. Solving Eq. \ref{eq:Volume} for $r_{i_{ss}}$ we will get:

\begin{equation}
\label{eq:InnerRadius}
r_{i_{ss}} = \pm \frac{\sqrt{(r_o\theta_{ss} + L_0)(r_o^3 \theta_{ss} + L_0 r_{i_0}^2)}}{r_o\theta_{ss} + L_0}
\end{equation}

\noindent where only the positive solution is acceptable. Considering Eq. \ref{eq:TorqueEquilibrium} and \ref{eq:InnerRadius} we can write:

\begin{equation}
\label{eq:BendingAngleSS}
\theta_{ss} = \frac{r_{hyd}\pi (\frac{\sqrt{(r_o\theta_{ss} + L_0)(r_o^3 \theta_{ss} + L_0 r_{i_0}^2)}}{r_o\theta_{ss} + L_0})^2 p_{hyd}}{k_{ss}}
\end{equation}

\noindent which is a second order equation of $\theta_{ss}$, the bending angle in steady states, and can be readily solved (only the positive solution is acceptable). Eq. \ref{eq:BendingAngleSS}, obviously, presents the bending angle in the steady states given the known parameters from the physical characteristics of the actuator.

Based on the above explanations, it is now possible to derive the kinematics and dynamic model of the RFEA (please refer to the Appendix)


\section{Experimental Validation}

\begin{table*}[t]
\caption{Estimated parameters}
\label{tab:EstimatedParam}
\begin{center}
\begin{tabular}{l | l | l | l}
Parameter									& 	2-DoF			&	8-DoF			&	units		\\
\hline
Damper Coef. in Neg. Direction ($b_{0{neg}}$)			&	$3.7374e-08$	&	$0.00058815$	&	Nms/deg	\\
Damper Lin. Func. Slope in Neg. Direction ($m_{b_{neg}}$)		& 	$-3.0518e-05$	&	$-0.18966$		&	Ns/deg	\\
Damper Coef. in Pos. Direction ($b_{0{pos}}$)			& 	$2.6171e-05$	&	$0.00032813$	&	Nms/deg	\\
Damper Lin. Func. Slope in Pos. Direction ($m_{b_{pos}}$)		& 	$-0.0038529$	&	$-0.098948$		&	Ns/deg	\\
Spring Coef. ($k_0$)							&	$0.00075223$		&	$0.045257$	&	Nm/deg	\\
Spring Lin. Func. Slope ($m_k$)						& 	$-0.00029452$	&	$-14.101$		&	N/deg		\\
Hydraulic Torque Arm ($r_{hyd}$)					&	$0.0037125$	&	$0.0035644$	&	m	
\\\end{tabular}
\end{center}
\end{table*}

In this section, we verify the performance of the proposed model both in simulation and experimentally. We also investigate the impact of the number of degrees of freedom in the accuracy of the proposed RSDA multibody system. To do so, two different models have been studied :$1)$a $2$-DoF model, and $2)$ an $8$-DoF model.

\begin{figure}
    \centering
    \begin{subfigure}[t]{8 cm}
        \includegraphics[trim={0 0 0 0},clip, width=8.0 cm]{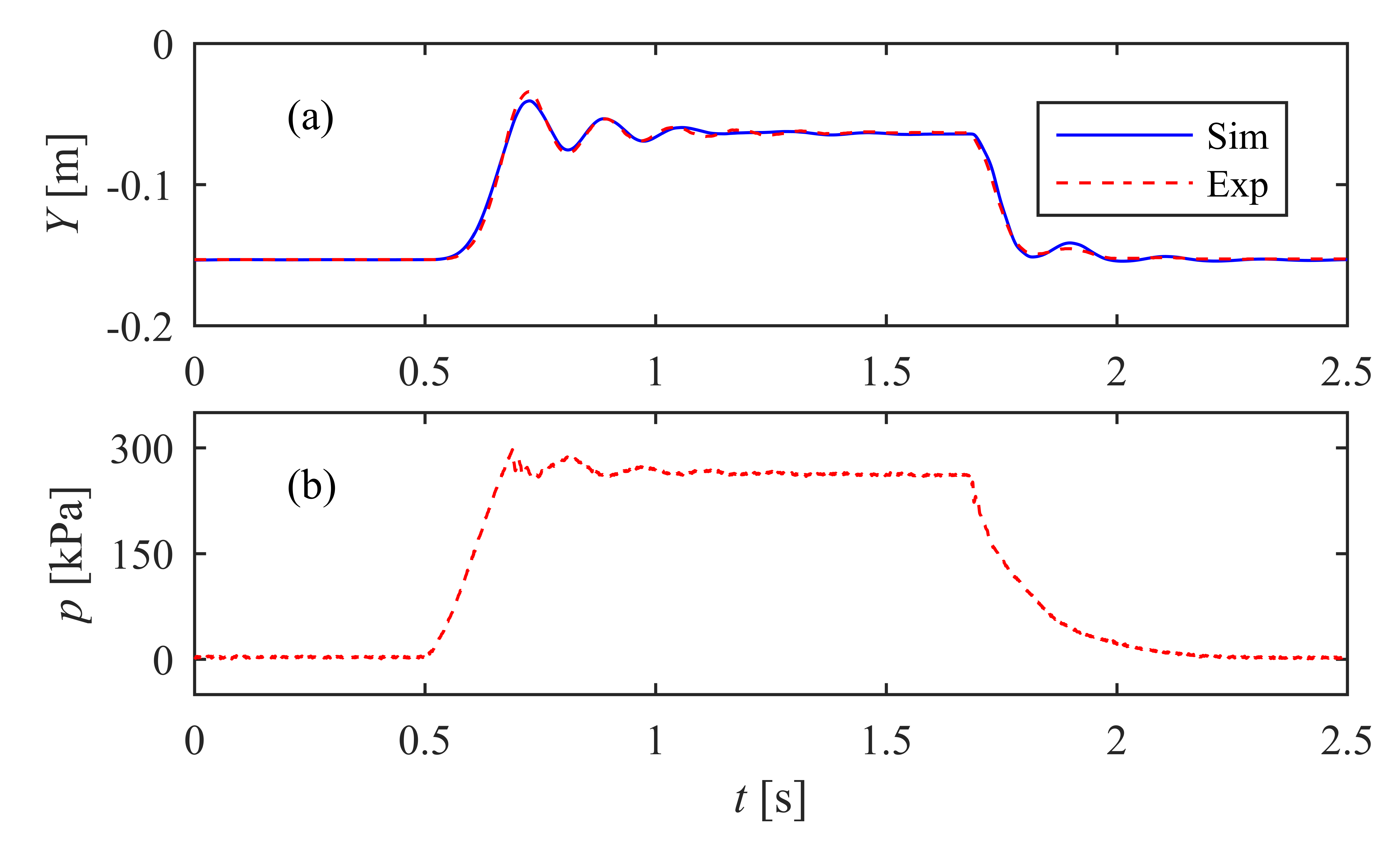}
        \caption{$2$-DoF model}
        \label{fig:EstimationExp2DOF}
    \end{subfigure}

    \begin{subfigure}[t]{8 cm}
        \includegraphics[trim={0 0 0 0},clip, width=8.0 cm]{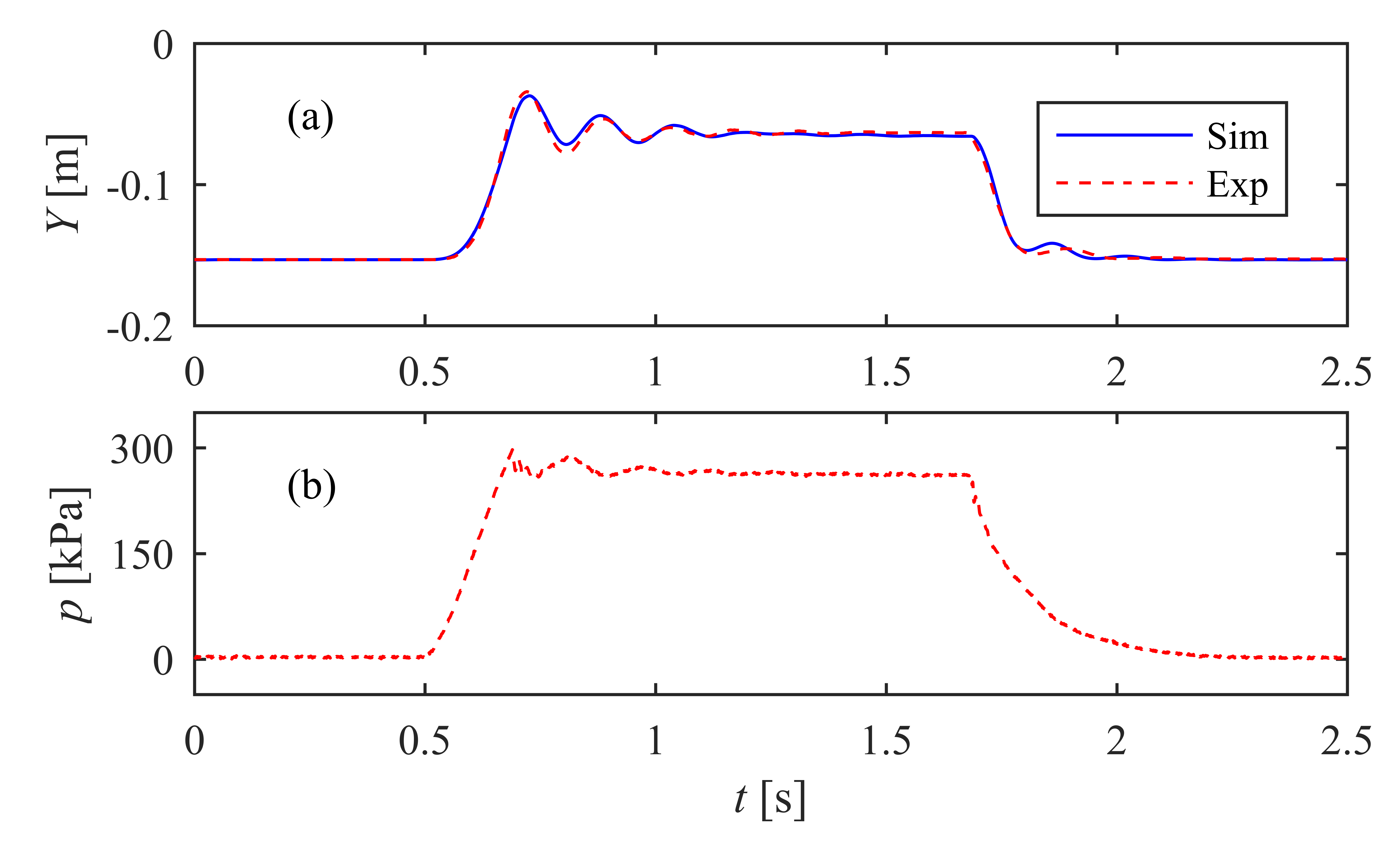}
        \caption{$8$-DoF model}
        \label{fig:EstimationExp8DOF}
    \end{subfigure}
    \caption{Experimental results and the fitted dynamic model. The tube is pressurized to $270kPa$.}
    \label{fig:DynModelEstimation}
\end{figure}

The current version of the fiber-reinforced elastomer manipulator used in this research work, Fig.~\ref{fig:ExpSetup}, is capable of carrying payloads up to $30g$. However, for verification and validation of the proposed model, the experiments are carried out while it is unloaded. The hydraulic drive system can accurately control the fluid pressure in real-time and move the tip to any position in its workspace. The position of the tip and the bending angle are measured using an image processing model implemented in MATLAB. The proposed $2$-DoF and $8$-DoF models of the system are also implemented in MATLAB. In both models, the inner radius of the tube, the spring coefficients and the damper coefficients were supposed to be variable and calculated dynamically as explained in Eq. \ref{eq:DynInnerRadius}, \ref{eq:SpringCoefficient} and \ref{eq:DamperCoefficient}. In order to verify the performance of the proposed model, the parameters of the model were identified. To do so, an independent experiment has been carried out while the soft actuator was pressurized until $270kPa$. The tip position and the hydraulic pressure have been recorded and the tip $y$ position has been used as a reference for parameter identification so that the simulation results  fit into the measured data. Simulink parameter estimation is used to determine the unknown model parameters shown in Table \ref{tab:EstimatedParam}. Initial assumptions for the parameter estimation are that the variables $b_{0_{neg}}$, $b_{0{pos}}$, $k_0$ are positive, whereas $0.003 m < r_{hyd} < 0.005 m$. Furthermore, it is assumed that the tube becomes less stiff when it elongates (inner radius increases); thus, $m_k$ has to be negative. Fig. \ref{fig:EstimationExp2DOF} and \ref{fig:EstimationExp8DOF} show the input fluid pressure and the corresponding tip $y$ position for both real system and the proposed $2$-DoF and $8$-DoF models, respectively. Table \ref{tab:EstimatedParam} also shows the identified parameters for both models.

\begin{figure}
    \centering
    \begin{subfigure}[t]{8 cm}
        \includegraphics[trim={0 0 0 0},clip, width=8.0 cm]{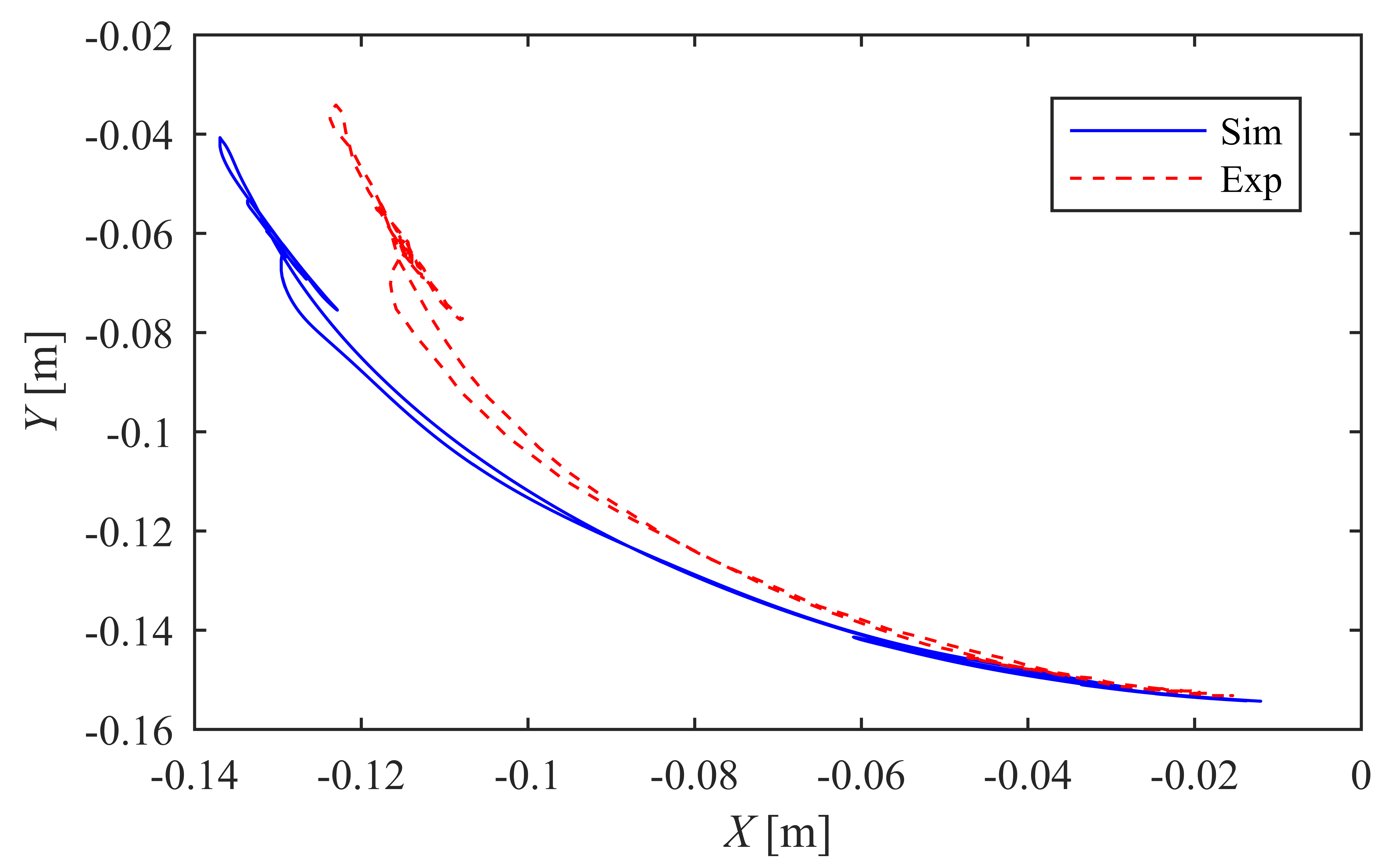}
        \caption{$2$-DoF model}
        \label{fig:EstimationExp2DOFPath}
    \end{subfigure}

    \begin{subfigure}[t]{8 cm}
        \includegraphics[trim={0 0 0 0},clip, width=8.0 cm]{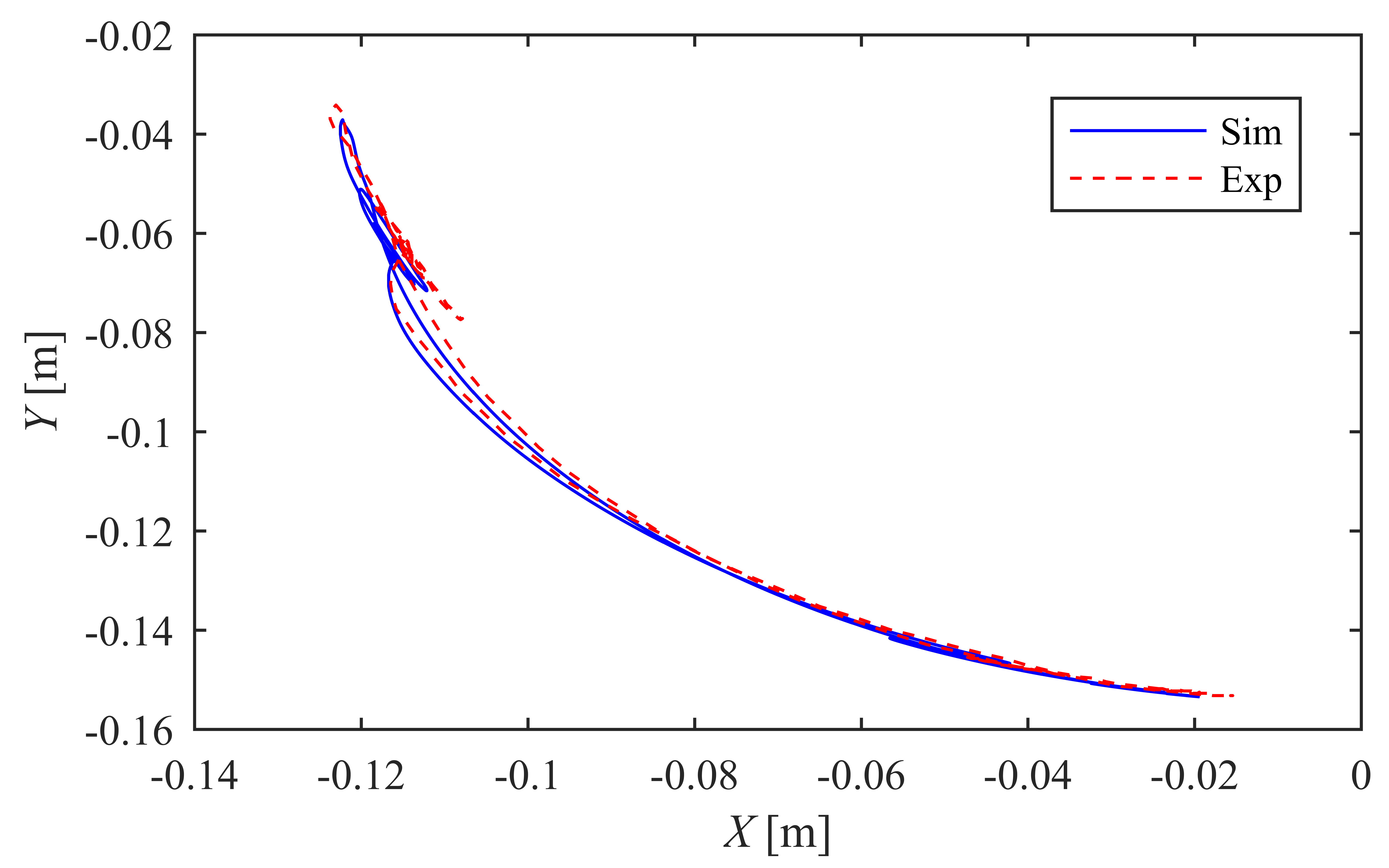}
        \caption{$8$-DoF model}
        \label{fig:EstimationExp8DOFPath}
    \end{subfigure}
    \caption{Path followed by actuator's tip when the tube is pressurized to $270kPa$.}
    \label{fig:DynModelEstimationPath}
\end{figure}

Even though the results of both $2$-DoF and $8$-DoF models look very promising at the first sight, their main difference appears when the $x$-coordinate is considered too.  Fig. \ref{fig:DynModelEstimationPath} shows the simulation results and the real path followed by the two models. Apparently, the lower the number of degrees of freedom, the higher discrepancy between the measured path and the simulated one. This was already obvious and predictable as the gravity will have different effect on the modeled system when the number of degrees of freedom changes.

\begin{figure}
    \centering
    \begin{subfigure}[t]{8 cm}
        \includegraphics[trim={0 0 0 0},clip, width=8.0 cm]{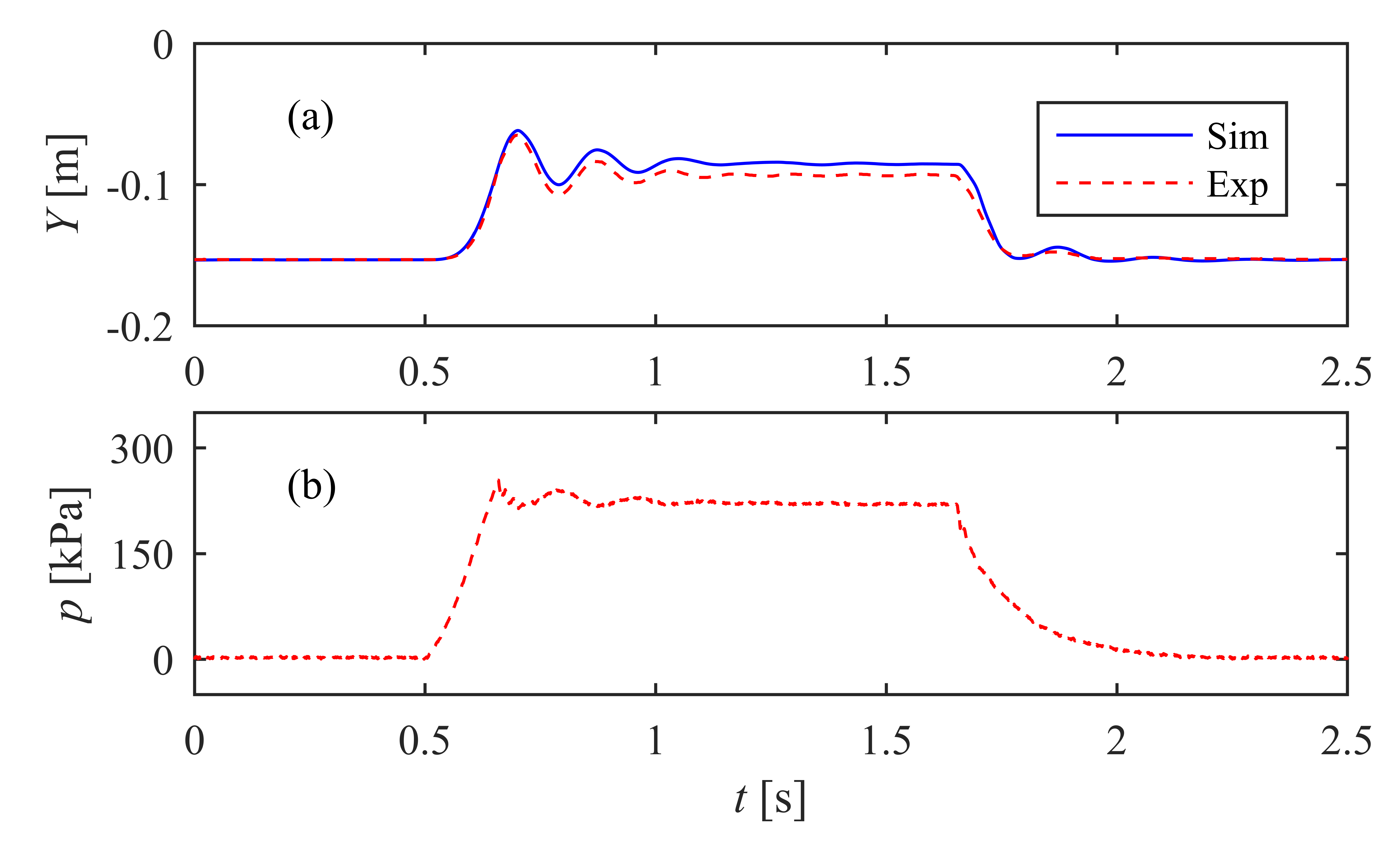}
        \caption{$2$-DoF model}
        \label{fig:DynModelValidation2DOF}
    \end{subfigure}

    \begin{subfigure}[t]{8 cm}
        \includegraphics[trim={0 0 0 0},clip, width=8.0 cm]{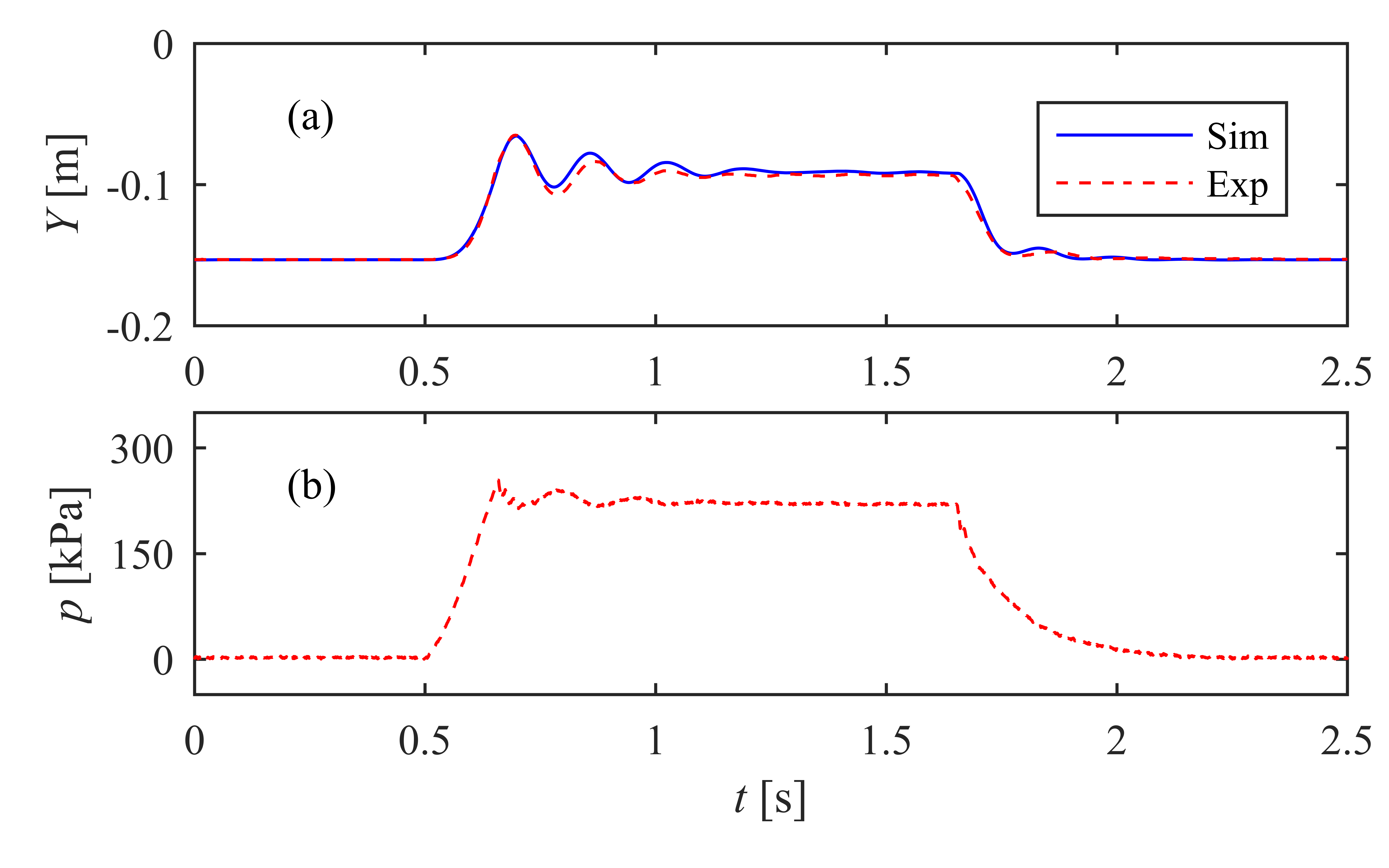}
        \caption{$8$-DoF model}
        \label{fig:DynModelValidation8DOF}
    \end{subfigure}
    \caption{Validation of the dynamic model. The tube is pressurized to $220kPa$.}
    \label{fig:DynModelValidation}
\end{figure}

\begin{figure}
    \centering
    \begin{subfigure}[t]{8 cm}
        \includegraphics[trim={0 0 0 0},clip, width=8.0 cm]{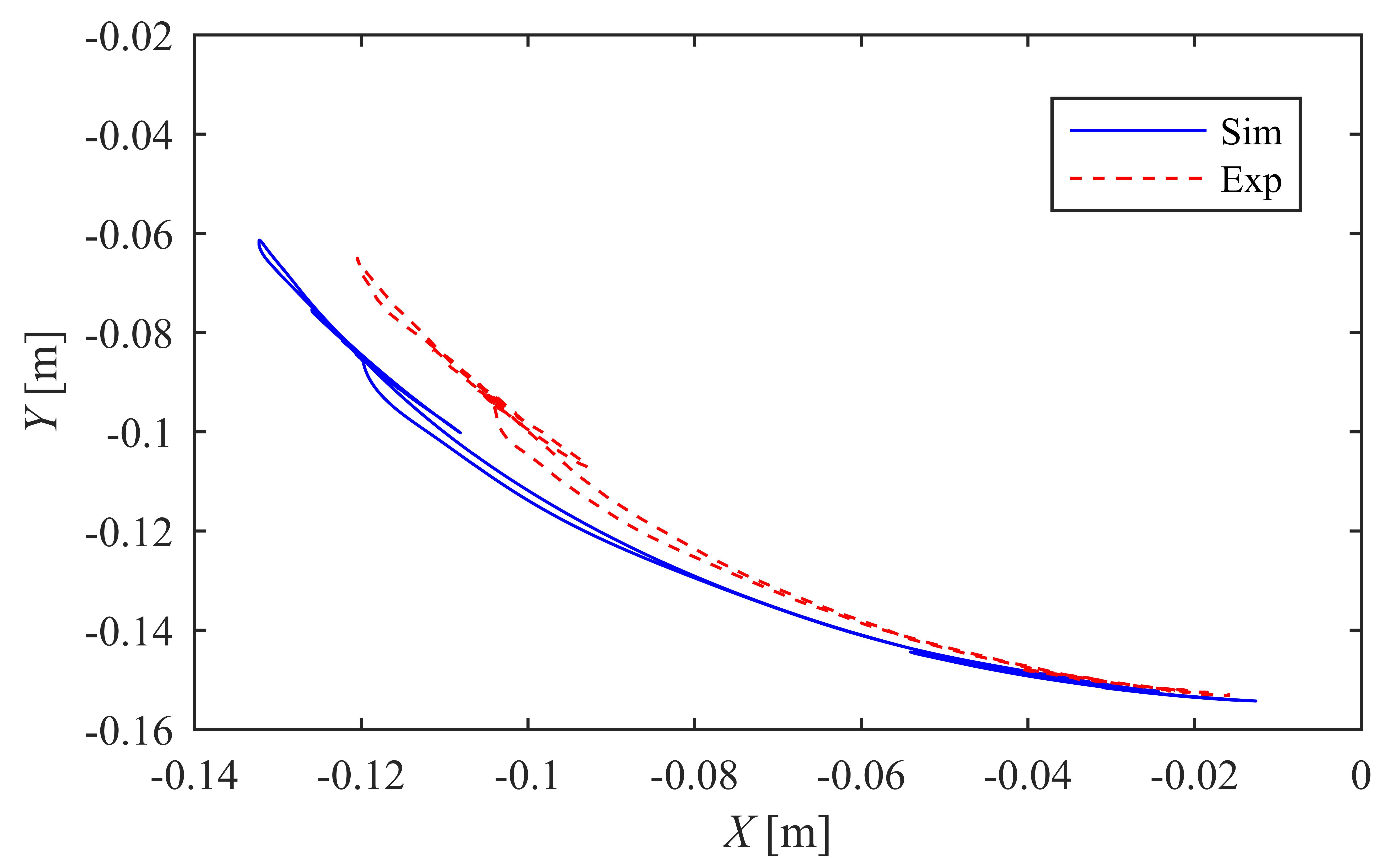}
        \caption{$2$-DoF model}
        \label{fig:DynModelValidation2DOFPath}
    \end{subfigure}

    \begin{subfigure}[t]{8 cm}
        \includegraphics[trim={0 0 0 0},clip, width=8.0 cm]{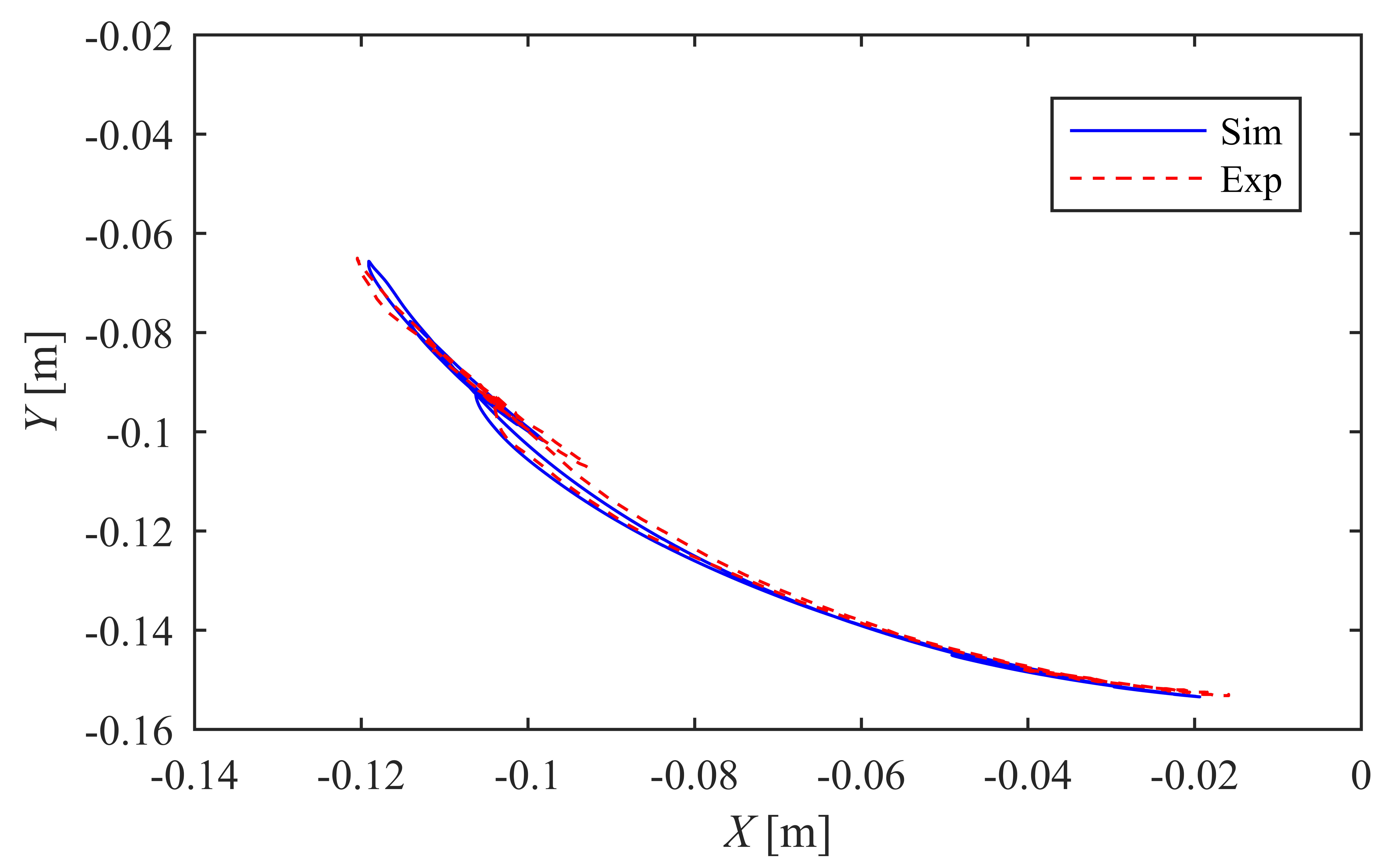}
        \caption{$8$-DoF model}
        \label{fig:DynModelValidation8DOFPath}
    \end{subfigure}
    \caption{Validation of path followed by actuator's tip when the tube is pressurized to $220kPa$.}
    \label{fig:DynModelValidationPath}
\end{figure}

To cross validate the performance of the proposed model, series of validation experiments have been done where different fluid pressure have been fed to the soft actuator and the tip positions were measured. The same fluid pressure have also been fed to the simulation model with the parameters previously identified and presented in table \ref{tab:EstimatedParam}. As an example, we pick the case where the tube is pressurized up to $220$ kPa. Fig. \ref{fig:DynModelValidation} shows the validation results for both $2$-DoF and $8$-DoF models in this case. Fig. \ref{fig:DynModelValidationPath} also shows the measured and simulated path followed by the tube's tip for $2$-DoF and $8$-DoF models, respectively.

\section{Discussion and Conclusion}

In this paper, a new modeling approach based on RSDA multibody system was proposed in order to model the soft material actuators. Two different models with $2$-DoF and $8$-DoF were experimentally investigated and the results were presented. As it can be seen, the $8$-DoF model outperforms the $2$-DoF significantly. Not only it reasonably accurately simulates the dynamic behavior of the soft actuator but also the path followed by the tip in the former model is quite similar to the measured one. Obviously, since the soft actuator is made of deformable material, higher number of degrees of freedom will provide us with more accurate results.  On the other hand, after some steps, increasing the number of degrees of freedom may not necessarily provide us with considerably more accurate results. So, there has to be a minimum number of RSDA bodies that satisfies the desired accuracy which its discussion is not in the scope of this paper. It should also be noted that in the current model, only the inner radius and the spring/damper coefficients were supposed to be variable. However, in reality, the mass of the system is also variable. Adding variable mass to the proposed model, more accurate results will be achieved. More importantly, our experiments show that the elongation of the tube is not uniform, i.e., some parts of the tube extends more than some other parts. This verifies our assumption that due to the material properties, the spring and damper coefficients are not identical for each segment of the tube. Furthermore, in our model we assumed that the coefficients are varying linearly. Under the assumption that they vary nonlinearly, the results could potentially be improved. Above all, not only it is possible to use our model for real-time control tasks, but also it is acceptably accurately covering/modeling the hysteresis behavior of the system.

Future work will include incorporating the variable mass and nonlinear coefficients to the model in addition to improving the hysteresis modeling of the system. We will also present the results when the model is used for real-time control tasks. Furthermore, we will also extend our models to support three-dimensional motion of such elastomer actuators and thus, allowing us to build digital models of complex robots made using these actuators.

\section{Acknowledgments}

The authors would like to thank Minna Poikelisp{\"a}{\"a} and Jarkko Per{\"a}l{\"a} for their great work in design and fabrication of the fiber-reinforced elastomer actuator used in this research work.

\bibliographystyle{ieeetr} 

\bibliography{ICRA18_Bib}


\section*{APPENDIX}

\subsection{Dynamic Model}

In order to model the dynamic behavior of the RFEA, Lagrange equation is used to find the dynamic model of the $n$ RSDA multibody system.

\subsubsection{Lagrangian Representation}

Lagrangian function $\mathbfcal{L}$  is defined as the difference between the kinetic energy and potential energy of the system:

\begin{equation}
\label{eq:Lagrangian}
\mathbfcal{L}(q,\dot{q}) = \mathbfcal{T}(q,\dot{q}) - \mathbfcal{U}(q)
\end{equation}

\noindent where $q$ is the general coordinates that completely locate the dynamic system. Knowing the Lagrangian, the equations of motion of the system can be calculated as follows:

\begin{equation}
\label{eq:LagrangianEoMTV}
\frac{d}{dt}\frac{\partial \mathbfcal{T}}{\partial \dot{q}} -  \frac{\partial \mathbfcal{T}}{\partial q} + \frac{\partial \mathbfcal{U}}{\partial q}= \tau
\end{equation}

\noindent where $\tau$ is the generalized force corresponding to the generalized coordinate $q$. In a general case, if $\mathbfcal{T}$ is an explicit function of time $t$, the time derivatives can be calculated symbolically as:

\begin{equation}
\label{eq:TimeDerivative}
\frac{d}{dt}\frac{\partial \mathbfcal{T}}{\partial \dot{q}}  = \frac{\partial}{\partial \dot{q}} (\frac{\partial \mathbfcal{T}}{\partial \dot{q}}) \ddot{q} + \frac{\partial}{\partial q}(\frac{\partial \mathbfcal{T}}{\partial \dot{q}}) \dot{q} + \frac{\partial}{\partial t}\frac{\partial \mathbfcal{T}}{\partial \dot{q}}
\end{equation}

\noindent and thus the equations of motion can be written as:

\begin{equation}
\label{eq:GeneralEoM}
\mathbfcal{M}(t, q)\ddot{q} - \mathbfcal{H}(t, q, \dot{q}) = \tau
\end{equation}

\noindent where:

\begin{equation}
\label{eq:MatrixMh}
\begin{cases}
\mathbfcal{M}(t, q) = \frac{\partial}{\partial \dot{q}} \frac{\partial \mathbfcal{T}}{\partial \dot{q}}\\
\mathbfcal{H}(t, q, \dot{q}) = -\frac{\partial}{\partial q}\frac{\partial \mathbfcal{T}}{\partial \dot{q}} \dot{q} - \frac{\partial}{\partial t}\frac{\partial \mathbfcal{T}}{\partial \dot{q}} + \frac{\partial \mathbfcal{T}}{\partial q} - \frac{\partial \mathbfcal{U}}{\partial q}
\end{cases}
\end{equation}\\

\subsubsection{Kinetic Energy}

Based on K{\"o}nig's theorem, the kinetic energy of a body is defined as:

\begin{equation}
\label{eq:KineticEnergy}
\mathbfcal{T}(q,\dot{q}) = \frac{1}{2}(m \mathbfcal{V}_{c}^T \mathbfcal{V}_{c} + \omega^T I \omega)
\end{equation}

\noindent where $m$, $\mathbfcal{V}_c$, $\omega$, and $I$ are the mass of each segment, translational velocity, angular velocity and moment of inertia of the body, respectively, in body frame. Note that the local coordinates are preferred here to avoid model complexity in hyper DoF.

\subsubsection{Potential Energy}

Resultant force acting on a multibody system can be represented as conservative and non-conservative forces. The former is given by partial derivatives of potential energy $\mathbfcal{U}$ in Lagrange equations of motion. Gravity and the spring force are the only conservative forces in the proposed modeling approach and thus, the total potential energy stored in the multibody system is given by:

\begin{equation}
\label{eq:PotentialEnergy}
\mathbfcal{U}(q) = \sum_{i=1}^n (\frac{1}{2}k_i \theta_i^2 -  m \mathbfcal{G}^T l_{c_i}(q) \cos(\sum_{j = 1}^i \theta_j))
\end{equation}

\noindent where $l_{c_i}(q)$, is the position vector of the center of mass of body $i$, which is a function of coordinate system, and $\mathbfcal{G}$ is the gravity. $\theta_i$'s reference are also the potential energy reference datum.

\subsubsection{Generalized Forces}

The generalized forces applied to the proposed $n$ RSDA  multibody system will include the effective torque applied by the fluid pressure difference to each corresponding body after damping:

\begin{equation}
\label{eq:GenForces}
\tau = \tau_{hyd} - \tau_d
\end{equation}

\subsubsection{Moment of Inertia}

It can be easily shown that the moment of inertia around the $x$-axis passing through the center of mass of each body in the proposed multibody system, a thick-wall cylindrical tube, can be calculated as:

\begin{equation}
\label{eq:CylinderIx}
I_{xx} = \frac{m(3({r_o}^2 + {r_i}^2) + \Delta L^2)}{12}
\end{equation}

Please note that in practice, inside of the tube is filled with fluidic medium and thus, when the medium moves, it should consume energy. Eq. \ref{eq:CylinderIx} is only an approximation which based on our experimental/simulation data, yields to an acceptably accurate result.

\subsection{Kinematics Model}

The tip position of the multibody system in Fig. \ref{fig:RSDAModel} can be easily calculated by the following equation:

\begin{equation}
\label{eq:KinematicsClosedForm}
\left[\begin{array}{c}
X \\
Y \\
\end{array}\right]
=
\left[\begin{array}{c}
-\sum_{i=1}^{n} \Delta L_i \sin(\sum_{j=1}^{i} \theta_j) \\
-\sum_{i=1}^{n} \Delta L_i \cos(\sum_{j=1}^{i} \theta_j) \\
\end{array}\right]
\end{equation}\\

\textit{Remark:} Two models with two different number of segments, $n_1\ne n_2$, can have different bending angles for the same tip $y$ positions. As a result, if someone tries to identify the parameters based on tip $y$ position, the bending angles could be different.


\end{document}